\documentclass[preprint,12pt]{elsarticle}

\usepackage[numbers]{natbib}
\usepackage{graphicx}  % add this in the preamble

\usepackage{siunitx}

\usepackage[table, dvipsnames]{xcolor}

\usepackage{needspace}

\usepackage{caption}   % For captions in tables
\usepackage{fix-cm}
\usepackage{graphicx,verbatim}
\usepackage[T1]{fontenc}
\usepackage[numbers]{natbib}
\usepackage{amsfonts}
\usepackage{algorithm}
\usepackage{algpseudocode}
\usepackage{colortbl}
\usepackage{tikz}
\usetikzlibrary{shapes.geometric}
\usepackage{dashrule}

\usepackage{standalone}

\usepackage[normalem]{ulem}
\usepackage{todonotes}

\newcommand{\cmark}{\checkmark}   % light tick ✓
\newcommand{\xmark}{\times}
\usepackage{hyperref}
\usepackage{pifont}

\usetikzlibrary{arrows.meta}
\usetikzlibrary{tikzmark,calc}
\usetikzlibrary{shapes,arrows}
\usetikzlibrary{spy}
\usetikzlibrary{positioning}
\usetikzlibrary{backgrounds}
\usetikzlibrary{decorations}
\usepackage{hyphenat}
\usepackage{multirow}
\usepackage{stfloats}
\usepackage{float}
\usepackage{cuted}
\usepackage{float}
 \usepackage{booktabs}
\newcommand\mybox[2][]{\tikz[overlay]\node[fill=blue!20,inner sep=2pt, anchor=text, rectangle, rounded corners=1mm,#1] {#2};\phantom{#2}}
\newcommand{\customsize}[2]{\fontsize{#1}{#2}\selectfont}

\usepackage{subcaption}
\usepackage{enumitem}
\newlist{enumerate*}{enumerate}{1}
\setlist[enumerate*]{label=(\alph*), itemjoin={{; }}, itemjoin*={{; and }}, afterlabel=}

\def\tsc#1{\csdef{#1}{\textsc{\lowercase{#1}}\xspace}}
\tsc{WGM}
\tsc{QE}
\usepackage{amssymb}
\usepackage[most]{tcolorbox}
\usepackage{colortbl}
\usepackage{amsmath}
\begin{document}

\begin{frontmatter}



\title{Differential-UMamba: Rethinking Tumor
Segmentation  Under Limited Data Scenarios}

 
\author[1]{Dhruv Jain\corref{cor1}}%[<options>]

\cortext[cor1]{Corresponding author}

\ead{dhruv.jain@insa-rouen.fr}
% \credit{Conceptualization of this study, Methodology, Software}



\author[3]{Romain Modzelewski}%[]


\author[2]{Romain Hérault}%[]

\author[1]{Clement Chatelain}%[]

\author[4]{Eva Torfeh}%[]

\author[4]{Sebastien Thureau}%[]

\affiliation[1]{Normandie Univ., INSA Rouen, LITIS, 76801, Saint Etienne du Rouvray, France}
\affiliation[2]{Normandie Univ., UNICAEN, ENSICAEN, CNRS, GREYC, 14000 Caen, France}
\affiliation[3]{Nuclear Medicine Department, Henri Becquerel Cancer Center and AIMS-Quantif Laboratory, Rouen, France}
\affiliation[4]{Radiotherapy Department, Henri Becquerel Cancer Center and AIMS-Quantif Laboratory, Rouen, France}

%% Abstract
\begin{abstract}
In data-scarce scenarios, deep learning models often overfit to noise and irrelevant patterns, which limits their ability to generalize to unseen samples. To address these challenges in medical image segmentation, we introduce Diff-UMamba, a novel architecture that combines the UNet framework with the mamba mechanism to model long-range dependencies. At the heart of Diff-UMamba is a noise reduction module, which employs a signal differencing strategy to suppress noisy or irrelevant activations within the encoder. This encourages the model to filter out spurious features and enhance task-relevant representations, thereby improving its focus on clinically significant regions. As a result, the architecture achieves improved segmentation accuracy and robustness, particularly in low-data settings.
Diff-UMamba is evaluated on multiple public datasets, including medical segmentation decathalon dataset (lung and pancreas) and AIIB23, demonstrating consistent performance gains of 1–3\% over baseline methods in various segmentation tasks. To further assess performance under limited data conditions, additional experiments are conducted on the BraTS-21 dataset by varying the proportion of available training samples. The approach is also validated on a small internal non-small cell lung cancer dataset for the segmentation of gross tumor volume in cone beam CT, where it achieves a 4-5\% improvement over baseline.

\noindent \textcolor{red}{Code will be made available upon publication.}

\end{abstract}

%\nocite{*}

% Keywords

\end{frontmatter}

\section{Introduction}
\label{sec:introduction}
Deep learning has revolutionized medical image segmentation, with convolutional neural networks (CNNs) providing state-of-the-art performance in a wide range of clinical applications~\cite{mienyeDeepConvolutionalNeural2025}. Most architectures for 3D medical image segmentation are based on the foundational UNet design. 
However, CNNs inherently struggle to model long-range dependencies due to their limited receptive field, making them less effective in segmenting complex anatomical structures that are diffused, irregular, or spread over multiple slices on volumetric scans~\cite{valanarasuMedicalTransformerGated2021, gaoMedicalImageSegmentation2025}.

To address this, sequence-based models, such as transformers~\cite{transformer} and mamba~\cite{mamba} have gained attention for their ability to capture global context through attention mechanisms and state-space representations~\cite{shamshadTransformersMedicalImaging2023}. 
 Sequence models are specifically designed to learn both local and global dependencies, allowing them to construct context-aware representations. This capability is particularly advantageous for segmentation tasks, where accurate boundary delineation depends on isolated features along with the continuity and structure embedded across the sequence. 
 \subsection{Background on Sequence Models}
The transformer~\cite{transformer} is a significant improvement over traditional recurrent neural networks (RNN \cite{rnn}) and long-short-term memory networks (LSTM \cite{lstm}) by replacing their inherently sequential processing with a fully parallel multihead attention mechanism. This operation is formally defined in Equation~\ref{eq:attention}:

\begin{equation}
\text{Attn}(Q, K, V) = \text{softmax}\left(\frac{Q K^\top}{\sqrt{d_k}}\right) V
\label{eq:attention}
\end{equation}

\noindent where \(Q, K, V\) denote the query, key, and value, respectively, and \(d_k\) is the dimensions of the query.
Another variant known as the Differential Transformer~\cite{yeDifferentialTransformer2024}, reduces contextual noise by removing common-mode signals by signal differencing. This differential attention mechanism is expressed in Equation~\ref{eq:differential_attention}:

\begin{equation}
\scalebox{0.9}{$
\operatorname{DiffAttn}(X) = \left(\operatorname{softmax}\left(\frac{Q_1 K_1^\top}{\sqrt{d}}\right) - \lambda \operatorname{softmax}\left(\frac{Q_2 K_2^\top}{\sqrt{d}}\right)\right) V
$}
\label{eq:differential_attention}
\end{equation}

\noindent where \(Q_1, K_1\) represent the query and key matrices for the primary attention that captures the relevant signal components, while \(Q_2, K_2\) correspond to the query and key used to estimate and subtract common-mode noise patterns. The parameter \(\lambda\) controls the degree of noise suppression.
Transformer-based architectures have shown promise in 3D segmentation. UNETR~\cite{hatamizadehUNETRTransformers3D2021} uses a vision transformer (ViT) encoder on non-overlapping 3D patches. The features of multiple layers of the transformer are fused into a convolutional decoder. However, UNETR suffers from high computational cost, lack of hierarchical representation, and lack of local inductive biases.
SwinUNETR~\cite{hatamizadehSwinUNETRSwin2022} addresses these issues using a hierarchical swin transformer~\cite{liuSwinTransformerHierarchical2021a}. It restricts self-attention to local windows and introduces shifted windows for cross-window interaction.

Mamba~\cite{mamba} is another sequential model which introduces a family of selective state space models (SSMs) that achieve \(\mathcal{O}(T)\) time and memory complexity while preserving the long-range context modeling capability of transformers. It processes 1D input sequences \(x(t) \in \mathbb{R}\) into outputs \(y(t) \in \mathbb{R}\) via a hidden state \(h(t) \in \mathbb{R}^N\), capturing long-term dependencies through linear ordinary differential equations (ODEs):

\begin{equation}
h'(t) = A h(t) + B x(t), \quad y(t) = C h(t),
\end{equation}

\noindent where \(A \in \mathbb{R}^{N \times N}\) is the state transition matrix and \(B \in \mathbb{R}^{N \times 1}\), \(C \in \mathbb{R}^{1 \times N}\) are projection matrices. Using a time scale \(\Delta\), continuous values are discretized using zero-order hold (ZOH):

\begin{equation}
\overline{A} = \exp(\Delta A), \quad \overline{B} = (\Delta A)^{-1}(\exp(\Delta A) - I) \cdot \Delta B.
\end{equation}
\noindent During training, mamba leverages global convolution to process entire sequences efficiently and in parallel. The convolutional kernel is defined as

\begin{equation}
\overline{K} = \left(C \overline{B}, C \overline{A} \overline{B}, \ldots, C \overline{A}^{M-1} \overline{B}\right),
\end{equation}

\noindent where \(M\) is the input length. This structured kernel \mbox{$\overline{K} \in \mathbb{R}^M$} enables efficient modeling of long-range dependencies via linear transformations.
UMamba~\cite{maUMambaEnhancingLongrange2024} incorporated state-space mamba blocks into a UNet for efficient 3D medical segmentation. They proposed two variants, UMamba-Bot (baseline) with mamba in the bottleneck and UMamba-Enc using a mamba block in each encoder layer. UMamba reduces computational costs while maintaining or improving performance compared to transformer-based models, making it suitable for resource-limited settings.

\subsection{Limitations of sequence models in medical imaging}
 However, the success of sequence models depends on the access to large-scale annotated datasets($>800$ volumes), which are scarce in the medical domain due to the high cost and effort involved in generating pixel-level labels~\cite{litjensSurveyDeepLearning2017a, liuFewshotLearningInference2024}.  
This data scarcity poses a major challenge: when trained on small datasets, sequence-based models tend to overfit, learning spurious noise patterns instead of clinically relevant features~\cite{diagnostics14131328, liOSLNetDeepSmallSample2020, powerGrokkingGeneralizationOverfitting2022, shaoTransformersMeetSmall2022}. 
To address these limitations, recent work has explored various strategies for adapting sequential models to data-constrained settings. Swin-UMamba~\cite{swinumamba} introduced a hybrid architecture that leverages pre-training on large-scale datasets such as ImageNet to transfer general visual priors to smaller target 2D domains, effectively reducing overfitting. 
However, there are fewer 3D networks specifically designed to address the challenges of small datasets.

\subsection{Our Contribution}
We propose an architecture that incorporates a dedicated module designed to learn and suppress noise patterns in the latent space in an unguided manner. This module is jointly trained with the main UNet backbone and helps the model focus on task-relevant features by reducing spurious activations. 
 The contributions of this paper are as follows.
\begin{itemize} 
\item We propose a Differential UMamba architecture for tumor segmentation which integrates a novel noise reduction module (NRM) to mitigate overfitting on small-scale datasets. To the best of our knowledge, this is the first differential network introduced for medical image segmentation that incorporates additional parameters that are designed to suppress noise. 

\item We analyze the latent space of models trained on different dataset sizes and show that overfitting creates patterns similar to those caused by injected noise. 

\item This method was extensively evaluated on multiple small-scale datasets, where it consistently demonstrated improved performance over existing state-of-the-art approaches.

\item A deep learning-based pipeline for tumor contour propagation was implemented, enabling the model to take advantage of prior contour information and improve segmentation performance. This method was adopted from a previous method for segmenting organs at risk(OAR)~\cite{LinMa2021}.

\end{itemize}
 \noindent 

The paper is organized as follows. In Section~\ref{sec:proposed}, we describe the datasets, the proposed method along with a study on the latent space. Section~\ref{sec:exp} presents the implementation details along with both quantitative and qualitative results. Finally, in Section~\ref{sec:limitation}, we discuss limitations and potential directions for future work.

\section{Materials and methods}

\label{sec:proposed}
\subsection{Datasets}
\begin{table}[t]
\centering
\setlength{\tabcolsep}{5pt} 
\caption{Summary of datasets used in our study.}
\resizebox{0.95\linewidth}{!}{
\begin{tabular}{lccc}
\hline
\textbf{Dataset} & \textbf{Modality} & \textbf{Targets} & \textbf{Number of Volumes} \\ \hline

MSD Task06-Lung & CT & Tumor & 63 \\

MSD Task07-Pancreas & Contrast-enhanced CT & Pancreas \& Tumor & 281 \\
Internal Dataset & Planning CT \& CBCT & Gross Tumor Volume & 564 \\
BraTS & MRI (T1, T1ce, T2, FLAIR) & Enhancing tumor, edema, necrosis & 1251 \\

AIIB23 & HRCT & Airways (no tumor labels) & 120 \\
\hline
\end{tabular}}
\label{tab:dataset_summary}
\end{table}

We performed experiments on three tumor-focused datasets: MSD~\cite{msd} (lung and pancreas), \mbox{BraTS-21}~\cite{brats1,brats2, brats3}, and an internal dataset, as well as a dataset for airway segmentation (AIIB23~\cite{aiib23}). The details of each dataset are provided in this section, with a summary presented in Table~\ref{tab:dataset_summary}.

\subsubsection{Medical Segmentation Decathlon Dataset}
The medical segmentation decathlon~\cite{msd} (MSD) challenge provides a comprehensive collection of 3D medical imaging datasets in various anatomical sites, with the aim of promoting the development of generalizable segmentation algorithms. 

\noindent \textbf{Lung Segmentation:}
This task involves segmenting non-small cell lung cancer (NSCLC) lesions from thoracic CT scans. The dataset consists of 63 annotated 3D CT volumes released as part of the medical segmentation decathlon dataset~\cite{msd} (Task06-Lung).
A single foreground label is used to identify the tumor regions. The task presents a significant challenge due to the wide heterogeneity in tumor morphology and density, as well as the presence of confounding anatomical structures within the thoracic cavity. We divided the dataset into 50 volumes for training and 13 volumes for testing.

\noindent \textbf{Pancreas Segmentation:}
 This task (Task07-Pancreas) focuses on the segmentation of the pancreas and pancreatic tumors from contrast-enhanced abdominal CT volumes. The dataset includes annotated 3D scans, each containing two target structures: the pancreas and any existing tumor lesions. The complexity of this task arises from the small size of the pancreas, its irregular shape, and its low contrast with the surrounding abdominal organs. In addition, the often sparse and subtle appearance of tumors introduces further challenges, particularly in handling severe class imbalance during training. The dataset consists of 281 contrast-enhanced abdominal CT scans with manual annotations of the pancreas and pancreatic tumors. The dataset was divided into 250 volumes for training and 31 volumes for testing.

\subsubsection{BRaTS-21 Dataset}
This dataset~\cite{brats1, brats2, brats3} focuses primarily on gliomas segmentation, which are among the most common and aggressive brain tumors.
It consists of 1,251 3D brain magnetic resonance imaging (MRI) scans, each containing four distinct imaging modalities. T1-weighted (T1), post-gadolinium T1-weighted (T1CE), T2-weighted (T2) and T2 fluid-attenuated inversion recovery (T2-FLAIR).
 They provide voxel-wise annotated ground-truth labels. The images have a fixed spatial resolution of \mbox{240 × 240 × 155} voxels and originate from multiple institutions using a variety of MRI scanners, ensuring diversity in acquisition settings.
The dataset includes segmentation labels for three key tumor subregions: \begin{enumerate*}
    \item Non-enhancing tumor core (NCR/NET)
    \item Enhancing tumor (ET)
    \item Peritumoral Edema (ED).
\end{enumerate*} 
These labels are used to optimize the network and are combined in the end to generate segmentation results for the whole tumor (WT), the tumor core (TC), and the enhanced tumor (ET). The dataset is randomly split, allocating 150 images to the test set, while the remaining images are used for training and validation.

\subsubsection{ Internal Dataset: GTV Segmentation }
This dataset was anonymized and privately held, consisting of data from 82 patients diagnosed with non-operated non-small cell lung cancer.
They received radiation therapy doses ranging from 60 to 70 Gy, all of whom gave their consent to the use of their data.
The imaging was performed with a \mbox{3D free-breathing} technique using iodine contrast, a Siemens CT scanner for planning, and a Varian CBCT scanner with a linac accelerator for imaging on board. Each patient typically had 6–7 CBCT scans and a planning CT, with sequential registration of the CBCT and the first aligned to the planning CT. For the label contours, the previous GTV ($GTV_{n-1}$) was applied to the current CBCT ($CBCT_n$), using a threshold of [-400, +175 HU] to isolate the GTV, followed by manual adjustments for significant anatomical changes. Data were divided into subsets with 61 patients assigned to the training set, 14 to the validation set, and the remaining 7 to the test set, ensuring that patient data did not overlap between sets. This corresponded to 476, 48, and 40 scans, respectively.

\subsubsection{AIIB23 Dataset}
The AIIB23 dataset~\cite{aiib23} was released as part of the challenge ``Airway-Informed Quantitative CT Imaging Biomarker for Fibrotic Lung Disease-2023'' at MICCAI. It was assembled to stimulate robust airway tree segmentation and prognostic biomarker discovery in patients with interstitial lung disease, where the distorted architecture of end‑stage fibrosis makes airway extraction unusually difficult.
Although tumor segmentations were not available, we included this dataset to enhance diversity and improve generalizability of our model.
The training partition in the challenge consists of 120 high‑resolution chest CT volumes acquired in the full-inspiration high-resolution clinical CT (HRCT) protocol from patients with progressive fibrotic interstitial lung disease, primarily idiopathic pulmonary fibrosis. All scans have  ($\leq 1$mm) slice spacing  and in‑plane pixel sizes around 0.7mm, reflecting routine HRCT. The dataset is randomly split, allocating 99 volumes to the training set, while the remaining volumes are used for the test set.

\subsection{Proposed Architecture}
\begin{figure}[t]
    \centering
    \includegraphics[width=\linewidth]{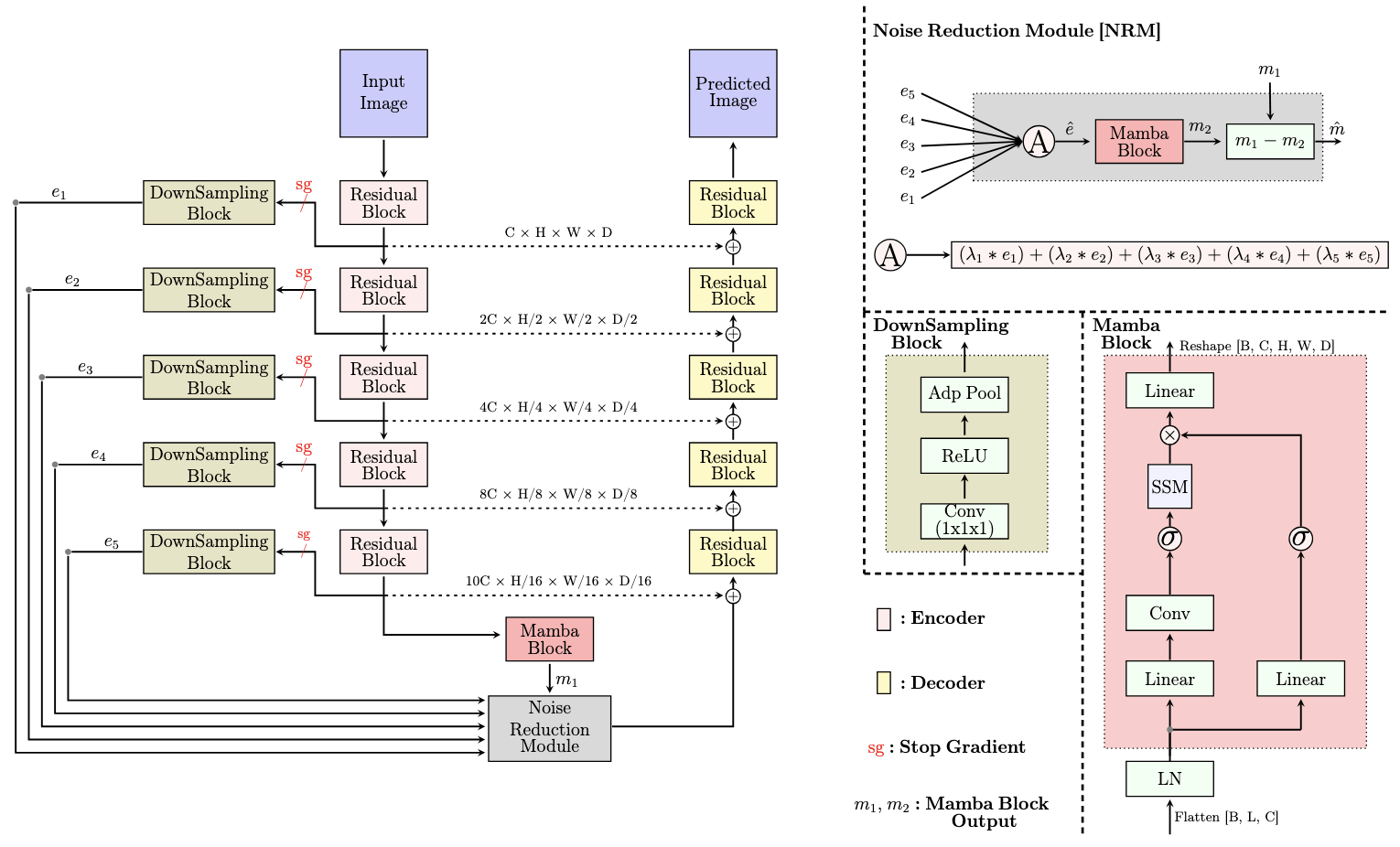}
    \caption{An overview of the proposed Diff-UMamba. Independent layers are employed to detect noise patterns, which is subsequently filtered through the noise reduction module \textbf{(NRM)}. (\mbox{$\lambda_1$-$\lambda_5$} are learnable parameters.) The feature shapes of $e_1$ to $e_5$, along with $m_1$ and $m_2$, are same as that of the last skip connection.}
    \label{fig:diffumamba}
\end{figure}

Small datasets, which lack sufficient data points for generalizability, can introduce unwanted noisy patterns into models due to overfitting or memorization of training \mbox{data \citep{alzubaidiSurveyDeepLearning2023}}. The proposed Diff-UMamba addresses this issue by integrating specialized blocks at each encoder layer, which learn the noise patterns in the data and ultimately filter them out in the bottleneck layer via a noise reduction module (NRM), as shown in Figure~\ref{fig:diffumamba}.
\noindent This method of filtering out unimportant features is inspired by the differential transformer~\cite{yeDifferentialTransformer2024}, which uses a negation technique in the attention layer to eliminate contextual token-level noise. 
The residual and mamba blocks of Diff-UMamba follow the same structure as used in UMamba~\cite{maUMambaEnhancingLongrange2024}. Each residual block consists of a convolutional layer, followed by instance normalization (IN~\cite{innorm}) and a Leaky ReLU~\cite{rectified} activation function. The original input is then added to the processed output, allowing the block to learn residual mappings.
The mamba block processes the input through two parallel paths. In both, the sequence is projected to a higher resolution of $(B, 2L, C)$ using linear layers. The first path applies a 1D convolution and a SiLU~\cite{silu} activation, along with a structured state-space (SSM) operation. The second applies another linear projection followed by SiLU~\cite{silu} activation. The outputs of both paths are combined element-wise using the hadamard product. Finally, the resulting features are projected back to $(B, L, C)$ and reshaped to the original 3D layout.
 Equation~\ref{eq:networkcomp} summarizes the complete forward pass of \mbox{Diff-UMamba}.
\begin{equation}
    \hat{y} = \mathcal{D} \left( \mathcal{M}_1(\mathcal{E}(x)) - \mathcal{M}_2\left( \sum_{i=1}^{l} \lambda_i e_i \right) \right)
    \label{eq:networkcomp}
\end{equation}
\noindent where $\hat{y}$ is the predicted output from the network, $\mathcal{E}$ is the primary encoder that extracts feature representations from input $x$, $\mathcal{D}$ is the decoder that reconstructs the final output, $\mathcal{M}_1$ is the bottleneck mamba block that
produces intermediate representation $m_1$, and $\mathcal{M}_2$ is the second mamba block that produces $m_2$ from a weighted sum of external embeddings $e_i$. 

\subsubsection{DownSampling Block}

Each of the outputs from the encoder layers is passed through a dedicated downsampling block to project the feature maps to a uniform spatial resolution. Each block consists of a $1 \times 1 \times 1$ convolution, a ReLU~\cite{rectified} activation, and an adaptive average pooling operation that resizes the feature maps to a predefined bottleneck size. Given an encoder output feature map $F_i \in \mathbb{R}^{C_i \times D_i \times H_i \times W_i}$ from the $i$-th encoder layer, the downsampled output $\hat{F}_i$ is computed as in Equation~\ref{eq:downsampling}:

\begin{equation}
\hat{F}_i = \text{AdapPool}\left( \text{ReLU}\left( \text{Conv}_{1\times1\times1}(F_i) \right) \right), \quad \hat{F}_i \in \mathbb{R}^{C' \times D' \times H' \times W'}
\label{eq:downsampling}
\end{equation}

\noindent Here, $C'$ is the target number of channels and $(D', H', W')$ represents the spatial desired dimensions of the bottleneck. Separate downsampling blocks are applied to each encoder layer output to ensure consistent dimensionality before feature fusion or aggregation in the NRM.

\subsubsection{Noise Reduction Module(NRM)}
This module is illustrated in Figure~\ref{fig:diffumamba}, with its inputs and outputs clearly depicted. It operates on the encoder-derived features \(e_1\) to \(e_5\), \(m_1\) and produces \(\hat{m}\) as output. The module aggregates noise patterns from multiple downsampling layers, as defined in Equation~\ref{eq:ensemble}:

\begin{equation}
\hat{e} = \sum_{i=1}^{l} \lambda_i e_i
\label{eq:ensemble}
\end{equation}

\noindent Here, $\lambda_i$ are trainable parameters and $e_i$ represent the output produced by the $i^{th}$ specialized downsampling block. After aggregating the outputs, they are processed by the mamba block $\mathcal{M}_2$ ~\cite{mamba} before being filtered out of the main bottleneck as shown in~Equation \ref{eq:m1m2}. 
\begin{equation}
   \hat m = m_1 - m_2
   \label{eq:m1m2}
\end{equation}
\noindent  where $m_1$ corresponds to the extracted features that contain useful information and noise, while $m_2$ represents the estimated noise. As in classical signal processing systems (e.g., Kalman filters), state-space models are known to perform signal denoising by estimating latent dynamics and filtering out measurement noise~\cite{kalman}. The use of mamba in the NRM is inspired by this principle. 
 Subtracting $m_2$ from $m_1$ effectively acts as a filter to isolate the meaningful features and suppress noise. 
 \begin{algorithm}[htbp]
\caption{NRM Integration into UNet-based architectures}
\label{alg:nrm-unet}
\begin{algorithmic}[1]
\Require Encoder outputs $\{F_1, F_2, \dots, F_l\}$, bottleneck feature $m_1$
\Require Trainable parameters $\{\lambda_1, \lambda_2, \dots, \lambda_l\}$, Mamba block $\mathcal{M}$
\Ensure Denoised output $\hat{m}$ for decoding

\For{$i = 1$ to $l$}
    \State $e_i \gets \text{AdapPool}(\text{ReLU}(\text{Conv}_{1\times1\times1}(F_i)))$ \Comment{Downsample $F_i$ to produce $e_i$ as in Equation~\ref{eq:downsampling} }
\EndFor

\State $\hat{e} \gets \sum\limits_{i=1}^{l} \lambda_i \cdot e_i$ \Comment{Aggregate encoder outputs using Equation~\eqref{eq:ensemble}}

\State $m_2 \gets \mathcal{M}(\hat{e})$ \Comment{Estimate noise using a Mamba block}

\State $\hat{m} \gets m_1 - m_2$ \Comment{Filter noise as per Equation~\eqref{eq:m1m2}}

\State \Return $\hat{m}$ \Comment{Pass $\hat{m}$ to decoder}
\end{algorithmic}
\end{algorithm}

\noindent The Noise Reduction Module (NRM) is a flexible component that can be integrated into any UNet-based architecture. The procedure for incorporating NRM into existing architectures is outlined in Algorithm~\ref{alg:nrm-unet}.
 \textbf{Note:} There are 2 mamba blocks in \mbox{Diff-UMamba}. We recommend this model for smaller datasets, where mamba models~\cite{mamba} (although they perform better than transformers~\cite{transformer}) often struggle to perform effectively. In contrast, larger datasets tend to offer a wider range of diverse training samples, allowing the model to generalize more effectively and generally perform better with sequence-based models. 

\subsection{Latent Space Analysis}

\begin{figure}[t]
\centering
\includegraphics[width=0.85\linewidth]{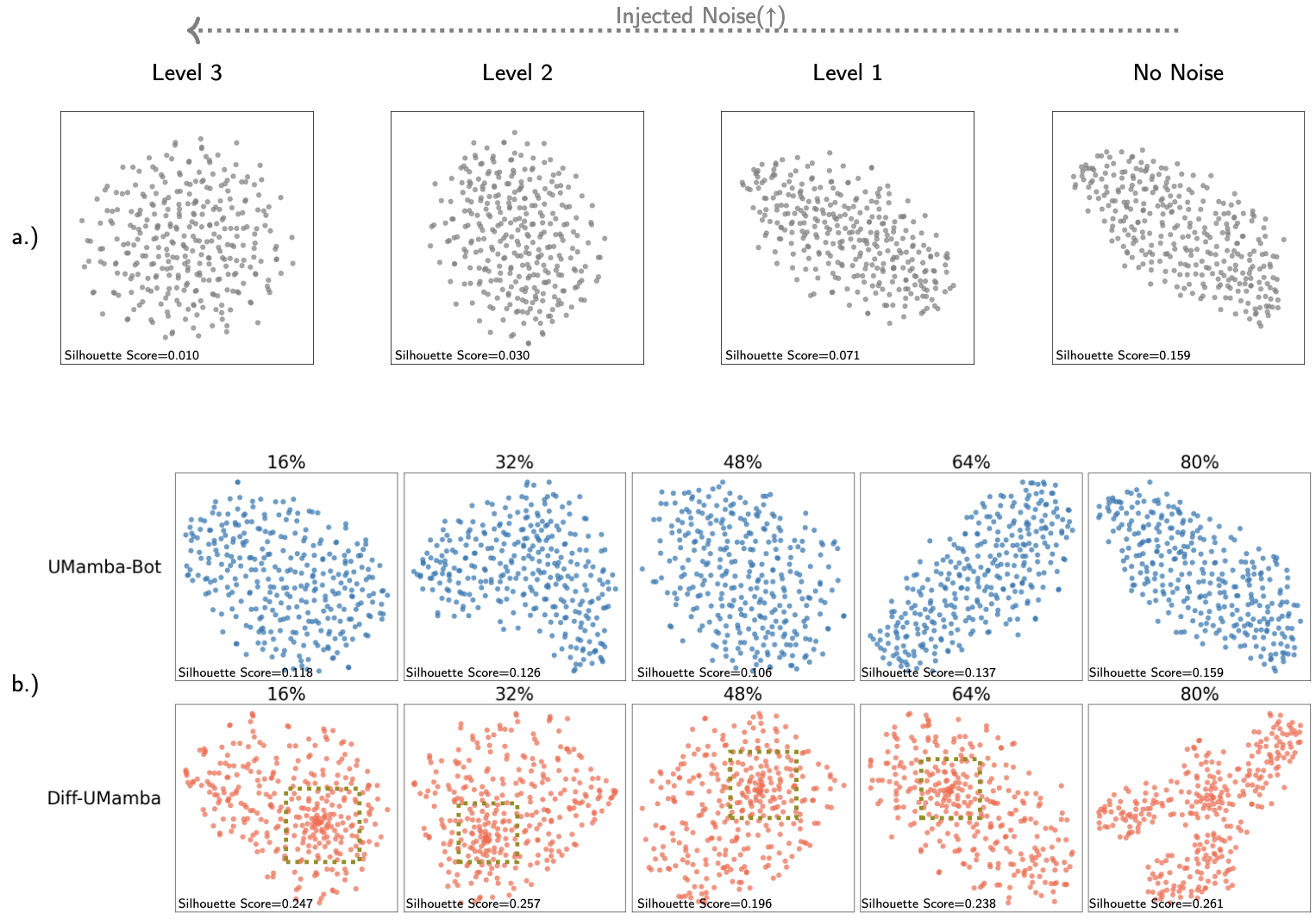}
        \caption{a.) Visualization of channel-wise feature shapes in the UMamba-Bot bottleneck under different levels of applied Gaussian noise. b.) t-SNE visualization of channel-wise bottleneck features for UMamba-Bot (top row) and Diff-UMamba (bottom row) across varying proportions (16\%, 32\%, 48\%, 64\%, 80\%) of the BRaTS-21 dataset. Each point represents a channel token, highlighting how feature representations evolve with increasing training data and model variation. }
    \label{fig:tsne}
\end{figure}

We found that models trained on limited datasets tend to capture spurious or overfitted patterns, which appear as noise-like distortions in the feature space. To explore this phenomenon, we visualized the bottleneck features of the trained model (UMamba-Bot~\cite{maUMambaEnhancingLongrange2024}). In Figure~\ref{fig:tsne}a, we show the channel-wise t-SNE projections, where varying levels of Gaussian noise were artificially added. As the noise level increased, the structured organization of the features progressively deteriorated, indicating the sensitivity of noise in the latent space.
Interestingly, a similar degradation in feature structure is observed when models are trained on reduced subsets of the dataset.
\noindent To demonstrate this, we visualize the bottleneck features of the models trained in the BraTS-21 dataset using different sizes of the training data in Figure~\ref{fig:tsne}b. As the size of the training set decreases, the feature embeddings become increasingly scattered and less coherent, suggesting that limited data induce noise-like artifacts in the latent space.
 To quantify this behavior, we employ the silhouette score~\cite{silhouette}, which for a data point $i$ is defined in Equation~\ref{eq:silhouette}:
 \begin{equation}
     s_i = \frac{b_i - a_i}{\max(a_i, b_i)}
     \label{eq:silhouette}
 \end{equation}
where $a_i$ denotes the average distance within the cluster (that is, the mean distance between $i$ and all other points in the same cluster), and $b_i$ represents the minimum average distance between $i$ and all points in the closest neighboring cluster.
Clustering is performed using k-means with clusters $k$, and maximum silhouette scores are used to evaluate the quality of clustering and select the optimal $k$. 
Reduced clustering behavior is observed between UMamba-Bot (16\%, 32\%, 48\%) which is similar to the increased levels of noise in the embedding space. 
However, analyzing the bottleneck features of the Diff-UMamba model in different training sample sizes, we observe that certain channels consistently exhibit well-separated clusters which are \textcolor{olive}{highlighted}.
Diff-UMamba have higher silhouette scores, suggesting more compact and discriminative feature representations. 
Although we cannot conclusively establish a direct causal link between clustered bottleneck characteristics and improvements in segmentation metrics (refer to Section~\ref{subsec:quanti}), we observe that this consistent clustering behavior appears in all Diff-UMamba models.

\subsubsection{Analysis of NRM}
\label{subsec:expanalysis}
In this section, we conduct exploratory experiments in the internal dataset to analyze the behavior of NRM and to compare the performance of Diff-UMamba with the baseline model.
\begin{figure}[ht]
    \centering
    \begin{subfigure}{0.40\textwidth}
        \centering
        \input{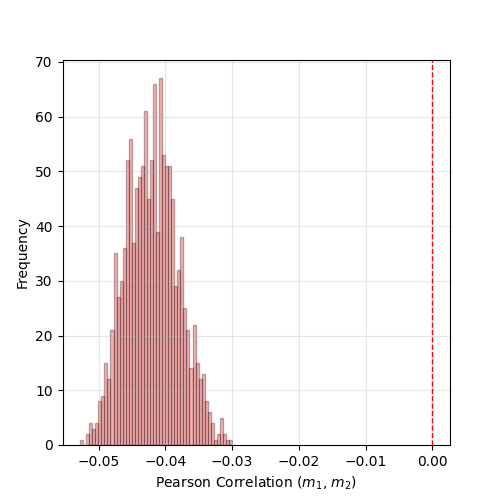}
        \caption{Pearson correlation between \mbox{$m_1$, $m_2$} embeddings.}
        \label{fig:lambdas}
    \end{subfigure}%
    % Space between subfigures 
    \hspace{0.05cm}
    \begin{tikzpicture}
        %\draw[dotted, thick] (0,1) -- (0,7.4); % Adjust height accordingly
    \end{tikzpicture}%
    \hspace{0.05cm}
    \begin{subfigure}{0.45\textwidth}
        \centering
        \input{noise_diagram}
        \caption{Effect of noise injection at the first residual block.
}
        \label{fig:noise}
    \end{subfigure}
    \caption{Insights into the NRM block}
    \label{fig:combined}
\end{figure}
\noindent Figure~\ref{fig:lambdas} illustrates the pearson correlation between $m_1$ and $m_2$ in 1,280 patches from the test set processed by the Diff-UMamba encoder. The results consistently show a low correlation between the two embedding types, indicating that they capture distinct feature representations. This suggests that $m_1$ and $m_2$ focus on different aspects of the input data.
Furthermore, we evaluated the robustness of the model by introducing various types of noise (gaussian, speckle, periodic, and salt-and-pepper) at different intensity levels into the first residual block during inference. As shown in Figure~\ref{fig:noise}, Diff-UMamba adapts more effectively than UMamba-Bot~\cite{maUMambaEnhancingLongrange2024}, demonstrating the ability of NRM to mitigate noise within the UMamba bottleneck. The mapping between noise levels and their corresponding parameters is given in the Appendix~\ref{fig:noise-mapping}.

\subsubsection{Evolution of lambda parameters}

\begin{figure}[h]
    \centering
    \begin{subfigure}{0.49\textwidth}
        \centering
        \includegraphics[width=\linewidth]{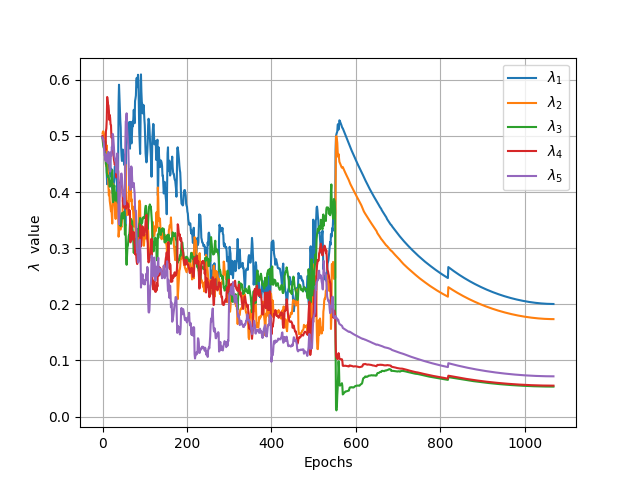}
        \caption{BRaTS-21 (32\%)}
        \label{fig:400lambda}
    \end{subfigure}%
    % Space between subfigures 
    \begin{subfigure}{0.49\textwidth}
        \centering
        \includegraphics[width=\linewidth]{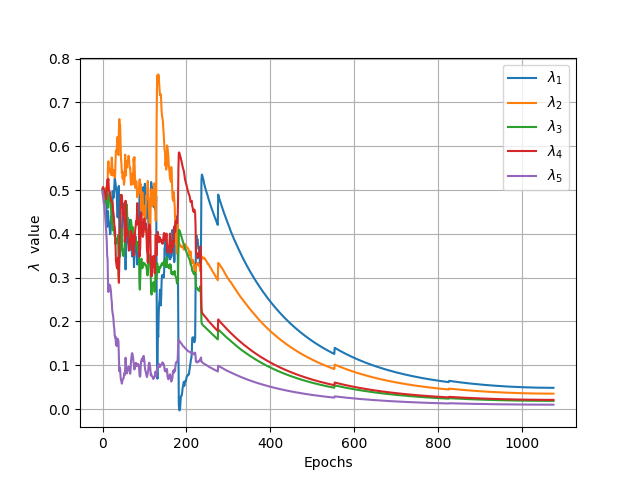}
        \caption{BRaTS-21 (80\%)}
        \label{fig:noise}
    \end{subfigure}
    \caption{Evolution of $\lambda$ during training for different number of samples. }
    \label{fig:lambda_evo}
\end{figure}

Figure~\ref{fig:lambda_evo} presents a comparison of the evolution of the $\lambda$-trainable parameters for models trained on 32\% and 80\% of the BRaTS-21~\cite{brats1, brats2, brats3} dataset. When trained on 32\% of the data, we observe that $\lambda_1$ and $\lambda_2$ attain significantly higher values than the remaining parameters. This suggests that, under limited data conditions, the model places greater importance on the earlier features (i.e., from shallower encoder stages).
\noindent In contrast, when trained on the 80\% of the dataset, the model assigns consistently lower values to all the parameters $\lambda$, with minimal variation between them. This indicates a reduced reliance on the noise-reducing effect of the differentiation operation. With access to sufficient data, the model can learn more expressive representations directly from the input without having to emphasize the differentiable signal. 
\noindent As a result, the need for the NRM diminishes, and the model implicitly down-regulates its influence by suppressing the values $\lambda$ during training. This behavior highlights the model’s ability to adaptively modulate its architectural mechanisms based on data availability.
Another observation is the stabilization behavior of the $\lambda$ parameters. In both settings 32\% and 80\% of the dataset, the $\lambda$ values initially exhibit fluctuations as the model searches for optimal combinations of noise features. 
However, stabilization occurs more quickly in the model trained on 80\% of the dataset.  The earlier stabilization supports the notion that the model has less reliance on feature differentiation when sufficient training samples are available, allowing it to settle more rapidly into an effective learned representation. 

\section{Experiments and results}
\label{sec:exp}

\subsection{Implementation}
We integrate Diff-UMamba into the UMamba-Bot architecture, which is built on the nnUNet~\cite{isenseeNnUNetSelfadaptingFramework2018} framework. It manages the selection of preprocessing, assembly, and network architecture, providing a robust and adaptable setup for medical segmentation tasks. Optimization was performed using SGD, with the unweighted sum of dice loss and cross-entropy loss. 
Experiments were conducted for 1000 epochs in the internal dataset, using a single input \mbox{(CBCT)} or two inputs (see Section~\ref{subsec:quanti}). For the public BRaTS-21 dataset~\cite{brats1, brats2, brats3}, four input modalities [T1, T1CE, T2, FLAIR] were used. In contrast, the MSD~\cite{msd} and AIIB23~\cite{aiib23} datasets utilized a single CT volume as input.
 All experiments were conducted with a Tesla A100 GPU.

\begin{figure}[t]
\noindent
\centering

\resizebox{0.75\linewidth}{!}{%
\begin{tikzpicture}[scale = 1]
\input{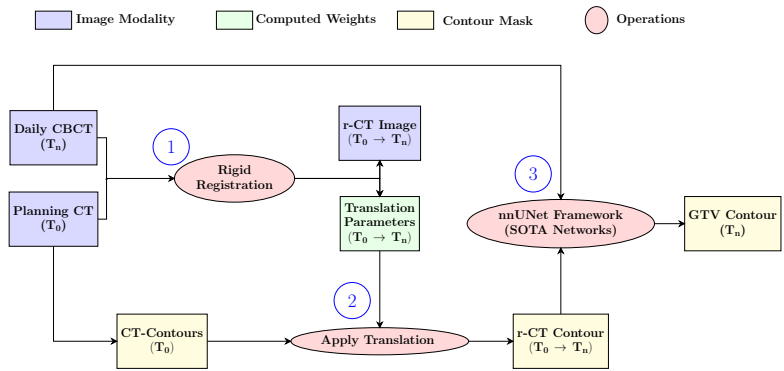}

\end{tikzpicture}
}
\caption{ Pipeline for segmenting GTV contours on CBCT. \mbox{\textcolor{blue}{1.) } CBCT} 
is rigidly-registered with Planning CT, \mbox{\textcolor{blue}{2.)} The} transformation parameters is used to transform the CT-contours to registered CT contour (r-CT contour), \mbox{\textcolor{blue}{3.)} CBCT} and \mbox{r-CT} contour are used as input to the corresponding model in the nnUNet framework~ \citep{isenseeNnUNetSelfadaptingFramework2018}. $T_0, T_n$ are the time points where the images were acquired. }
\label{fig:pipeline}
\end{figure}
\subsection{Quantitative Comparison with State-of-the-Art}
\label{subsec:quanti}
In this section, we evaluate and analyze various state-of-the-art segmentation networks across diverse datasets. 
Diff-UMamba is compared with various architectures such as CNN-based methods (nnUNetv2~\cite{isenseeNnUNetSelfadaptingFramework2018}, SegResNet~\cite{myronenko3DMRIBrain2018}), transformer-based methods (UNETR~\cite{hatamizadehUNETRTransformers3D2021}, SwinUNETR~\cite{hatamizadehSwinUNETRSwin2022}), and mamba-based models (LKM-UNet~\cite{lkmunet}, UMamba-Bot~\cite{maUMambaEnhancingLongrange2024}, UMamba-Enc~\cite{maUMambaEnhancingLongrange2024}, and SegMamba~\cite{segmamba}) for the internal dataset. 
Training a deep learning network for GTV segmentation presents significant challenges due to the variability in tumor shapes and sizes, as well as their frequent occurrence in low contrast regions~\citep{Liu2021}. 
To address this, we incorporate the rigidly registered contour (r-CT) as prior information to guide the network in accurately delineating the GTV as shown in Figure~\ref{fig:pipeline} similar to the approach demonstrated earlier for OAR segmentation in CBCT volumes~\citep{LinMa2021}.
The process begins with the registration of the planning CT (traditional radiotherapy timeline is shown in Figure~\ref{fig:trad-radio}) with the CBCT volume. 
\noindent The transformation parameters are optimized using gradient descent to align the two modalities. 
To minimize manual intervention, identical hyperparameters are used for all patients.
\textbf{Note:} This method is applied exclusively to the internal dataset.
\begin{figure}[h]
    \centering
        \centering
        \captionof{table}{
Comparison of state-of-the-art segmentation models with and without r-CT contours on the internal dataset for CBCT segmentation.(* indicates t-test p-value < 0.05; ** indicates p-value < 0.001)}
        \label{tab:private}
        
\setlength{\tabcolsep}{4pt}
\begin{tabular}{c|ccc|ccc|c}
\hline
\multirow{2}{*}{Models} & \multicolumn{3}{c|}{without r-CT} & \multicolumn{3}{c|}{with r-CT} & \multirow{2}{*}{Params(M)} \\
                        & DSC & IOU & HD95 & DSC & IOU & HD95 & \\ \hline
nnUNetv2~\cite{isenseeNnUNetSelfadaptingFramework2018}    
& 52.31* & 43.34 & 11.99 
& 71.04* & 61.31 & 4.20 & 30.10 \\

SegResNet~\cite{myronenko3DMRIBrain2018}                  
& 56.71* & 46.53 & 7.92 
& 72.73* & 63.50 & 4.07 & 18.80 \\

UNETR~\cite{hatamizadehUNETRTransformers3D2021}           
& 43.61** & 33.42 & 23.72 
& 69.56* & 59.61 & 4.24 & 121.40 \\

SwinUNETR~\cite{hatamizadehSwinUNETRSwin2022}             
& 54.67* & 42.68 & 10.34 
& 69.34* & 60.17 & 4.62 & 62.19 \\

SegMamba~\cite{segmamba}                                 
& 49.93** & 40.59 & 12.53 
& 71.73* & 61.50 & 3.95 & 66.87 \\

LKM-UNet~\cite{lkmunet}                                  
& 51.48* & 42.19 & 11.82 
& 72.64 & 62.47 & 3.48 & 63.95 \\

UMamba-Bot~\cite{maUMambaEnhancingLongrange2024}         
& 54.26* & 45.34 & 11.55 
& 72.47** & 61.85 & 3.83 & 42.12 \\

UMamba-Enc~\cite{maUMambaEnhancingLongrange2024}         
& 57.44 & 47.96 & 8.68 
& 74.63* & 63.38 & 3.19 & 42.86 \\

Diff-UMamba                                              
& \textbf{59.28} & \textbf{49.47} & \textbf{7.72} 
& \textbf{76.91} & \textbf{65.87} & \textbf{3.18} & 43.05 \\ \hline
\end{tabular}

\end{figure}
Table~\ref{tab:private} shows the results, divided into two groups: one excluding r-CT contours and the other including them in the input. Evaluation metrics consist of the dice coefficient (DSC~$\uparrow$), intersection over union (IOU~$\uparrow$), and hausdorff distance (HD95~$\downarrow$), where we observed that the incorporation of r-CT contours provides significant benefits to all models. Furthermore, \mbox{Diff-UMamba} gives better results in both cases than in the rest. The DSC increases by $5.02\%$ without \mbox{r-CT contours} and by $4.44\%$ with \mbox{r-CT contours}, compared to \mbox{Umamba-Bot}. Interestingly, the improvement is smaller when additional data (r-CT contour) is included.
The NRM increases the total parameter count by 2\%, while providing a noise reduction in the latent feature space.
\begin{table}[h]
\centering
\setlength{\tabcolsep}{5pt}
\centering
\caption{DSC comparison for contour propagation methods.}
\resizebox{0.6\linewidth}{!}{
\begin{tabular}{lccccc}
\cline{2-6}
                                 & Translation & Rigid & DIR-T & DIR-L & Ours  \\ \hline
\multicolumn{1}{c}{DSC} & 55.19                & 56.89          & 63.20          & 66.59          & \textbf{76.91}
\end{tabular}}
\label{fig:reg}
\end{table}
\begin{table}[htbp]
\caption{Comparison of segmentation performance on CT datasets: MSD~\cite{msd} and AIIB23~\cite{aiib23}. The table reports DSC, IoU, and HD95.}
\centering
\resizebox{\linewidth}{!}{%
\begin{tabular}{c|cccccc|ccc}
\hline
\multirow{3}{*}{Models} & \multicolumn{6}{c|}{Medical Segmentation Decathlon} & \multicolumn{3}{c}{AIIB23} \\ \cline{2-10} 
                        & \multicolumn{3}{c}{Lungs} & \multicolumn{3}{c|}{Pancreas(Avg)} & \multicolumn{3}{c}{Airway} \\ \cline{2-10}
                        & DSC($\uparrow$) & IOU($\uparrow$) & HD95($\downarrow$) & DSC($\uparrow$) & IOU($\uparrow$) & HD95($\downarrow$) & DSC($\uparrow$) & IOU($\uparrow$) & HD95($\downarrow$) \\ \hline
nnUNetv2~\cite{isenseeNnUNetSelfadaptingFramework2018} & 66.43 & 55.18 & 40.62 & 66.21 & 54.48 & 8.66 & 94.38 & 89.38 & 1.63 \\ 
SegResNet~\cite{myronenko3DMRIBrain2018} & 66.50 & 53.73 & 26.83 & 62.95 & 51.17 & 11.84 & 93.79 & 88.39 & 4.74 \\
UMamba-Bot~\cite{maUMambaEnhancingLongrange2024} & 71.28 & 58.78 & 23.87 & 66.47 & 54.41 & 8.97 & 93.48 & 88.02 & 7.47 \\ 
UMamba-Enc~\cite{maUMambaEnhancingLongrange2024} & 71.29 & 58.75 & 71.82 & 68.32 & 56.99 & \textbf{8.24} & 92.79 & 86.76 & 9.52 \\ 
LKM-UNet~\cite{lkmunet} & 67.61 & 56.66 & 27.68 & 67.70 & 55.62 & 10.73 & 94.09 & 88.90 & 1.97 \\
Diff-UMamba & \textbf{72.24} & \textbf{60.03} & \textbf{23.79} & \textbf{68.96} & \textbf{57.09} & 8.48 & \textbf{94.47} & \textbf{89.55} & \textbf{1.58} \\ \hline
\end{tabular}}
\label{tab:msd}
\end{table}

Table~\ref{fig:reg} presents a comparison of the non-deep learning contour propagation method that includes DIR with two variants: \mbox{DIR-T}, where only the tumor is deformed and \mbox{DIR-L}, where the entire lung is deformed. Our method consistently outperforms all other methods by a margin of at least 10\%. Figure~\ref{fig:dir} provides more details on the two variants of deformable registration.

We conducted experiments on the MSD~\cite{msd} dataset with tasks such as the lungs and pancreas along with the AIIB23~\cite{aiib23} dataset with airway segmentation. The results are shown in Table~\ref{tab:msd}. 
\begin{table}[h]
\centering
\caption{Comparison of state-of-the-art segmentation models on the BRaTS-21 dataset with varying proportion (16\%, 32\%, 48\%) of training samples. The table reports DSC, IoU, and HD95 as (mean $\pm$ std).   
}

\begin{tabular}{c|ccccccccc}
\hline
\multirow{3}{*}{Models} & \multicolumn{9}{c}{Avg (WT,  TC, ET)} \\ \cline{2-10}
& \multicolumn{3}{c|}{16\%} & \multicolumn{3}{c|}{32\%} & \multicolumn{3}{c}{48\%} \\ 
\arrayrulecolor{black}\cline{2-10}
& DSC($\uparrow$) & IOU($\uparrow$) & \multicolumn{1}{c|}{HD95($\downarrow$)} & DSC($\uparrow$) & IOU($\uparrow$) & \multicolumn{1}{c|}{HD95($\downarrow$)} & DSC($\uparrow$) & IOU($\uparrow$) & HD95($\downarrow$) \\
\hline
nnUNetv2~\cite{isenseeNnUNetSelfadaptingFramework2018}             
& 89.87{\scriptsize$\pm$15.11} & 83.97{\scriptsize$\pm$17.63} & \multicolumn{1}{c|}{5.00{\scriptsize$\pm$16.60}} 
& \textbf{91.43}{\scriptsize$\pm$12.70} & 85.96{\scriptsize$\pm$15.30} & \multicolumn{1}{c|}{4.36{\scriptsize$\pm$15.55}} 
& 91.25{\scriptsize$\pm$13.50} & 86.08{\scriptsize$\pm$15.47} & 3.61{\scriptsize$\pm$13.40} \\

SegResNet~\cite{myronenko3DMRIBrain2018}                
& 89.31{\scriptsize$\pm$16.73} & 83.90{\scriptsize$\pm$17.87} & \multicolumn{1}{c|}{4.95{\scriptsize$\pm$18.24}} 
& 90.74{\scriptsize$\pm$14.43} & 85.43{\scriptsize$\pm$16.29} & \multicolumn{1}{c|}{3.86{\scriptsize$\pm$14.16}} 
& 90.83{\scriptsize$\pm$14.47} & 85.39{\scriptsize$\pm$16.79} & 4.19{\scriptsize$\pm$15.18} \\

UMamba-Bot~\cite{maUMambaEnhancingLongrange2024}             
& 89.74{\scriptsize$\pm$15.38} & 84.03{\scriptsize$\pm$17.24} & \multicolumn{1}{c|}{5.01{\scriptsize$\pm$16.89}} 
& 90.59{\scriptsize$\pm$13.44} & 84.70{\scriptsize$\pm$15.79} & \multicolumn{1}{c|}{4.79{\scriptsize$\pm$16.10}} 
& 91.18{\scriptsize$\pm$13.74} & 86.21{\scriptsize$\pm$14.98} & 3.55{\scriptsize$\pm$11.01} \\

UMamba-Enc~\cite{maUMambaEnhancingLongrange2024}             
& 89.17{\scriptsize$\pm$16.12} & 83.28{\scriptsize$\pm$17.91} & \multicolumn{1}{c|}{5.16{\scriptsize$\pm$16.95}} 
& 90.54{\scriptsize$\pm$14.09} & 84.80{\scriptsize$\pm$16.54} & \multicolumn{1}{c|}{5.14{\scriptsize$\pm$16.95}} 
& 91.14{\scriptsize$\pm$13.65} & 86.14{\scriptsize$\pm$15.09} & 3.51{\scriptsize$\pm$10.78} \\

LKM-UNet~\cite{lkmunet}                 
& 89.23{\scriptsize$\pm$16.45} & 83.49{\scriptsize$\pm$18.24} & \multicolumn{1}{c|}{5.58{\scriptsize$\pm$18.63}} 
& 90.94{\scriptsize$\pm$13.60} & 85.34{\scriptsize$\pm$15.97} & \multicolumn{1}{c|}{4.63{\scriptsize$\pm$15.90}} 
& 90.77{\scriptsize$\pm$14.07} & 85.19{\scriptsize$\pm$16.59} & 4.64{\scriptsize$\pm$15.82} \\
Diff-UMamba             
& \textbf{89.93}{\scriptsize$\pm$\textbf{14.83}} & \textbf{84.21}{\scriptsize$\pm$\textbf{16.82}} & \multicolumn{1}{c|}{\textbf{4.85}{\scriptsize$\pm$\textbf{16.15}}} 
& 91.24{\scriptsize$\pm$\textbf{12.67}} & \textbf{86.22}{\scriptsize$\pm$\textbf{14.56}} & \multicolumn{1}{c|}{\textbf{3.41}{\scriptsize$\pm$\textbf{14.56}}} 
& \textbf{91.30}{\scriptsize$\pm$\textbf{13.10}} & \textbf{86.27}{\scriptsize$\pm$\textbf{14.51}} & \textbf{3.19}{\scriptsize$\pm$\textbf{9.78}} \\ 
\hline
\end{tabular}

\label{tab:brats}
\end{table}
\begin{table}[h]
\centering
\caption{Comparison of state-of-the-art segmentation models on the BRaTS-21 dataset with varying proportion  (64\%, 80\%) of training samples. The table reports DSC, IoU, and HD95 as (mean $\pm$ std).   
}

\begin{tabular}{c|cccccc}
\hline
\multirow{3}{*}{Models} & \multicolumn{6}{c}{Avg (WT,  TC, ET)} \\ \cline{2-7}
& \multicolumn{3}{c|}{64\%} & \multicolumn{3}{c}{80\%} \\ 
\arrayrulecolor{black}\cline{2-7}
& DSC($\uparrow$) & IOU($\uparrow$) & \multicolumn{1}{c|}{HD95($\downarrow$)} & DSC($\uparrow$) & IOU($\uparrow$) & HD95($\downarrow$) \\
\hline
nnUNetv2~\cite{isenseeNnUNetSelfadaptingFramework2018}             
& 91.64{\scriptsize$\pm$12.45} & 86.26{\scriptsize$\pm$15.10} & \multicolumn{1}{c|}{4.03{\scriptsize$\pm$15.67}} 
& 91.49{\scriptsize$\pm$13.14} & 86.16{\scriptsize$\pm$15.62} & 4.30{\scriptsize$\pm$15.48} \\

SegResNet~\cite{myronenko3DMRIBrain2018}                
& 91.05{\scriptsize$\pm$14.60} & 85.92{\scriptsize$\pm$16.40} & \multicolumn{1}{c|}{3.80{\scriptsize$\pm$14.40}} 
& 91.30{\scriptsize$\pm$13.88} & 86.26{\scriptsize$\pm$15.84} & 3.72{\scriptsize$\pm$14.42} \\

UMamba-Bot~\cite{maUMambaEnhancingLongrange2024}             
& \textbf{91.71}{\scriptsize$\pm$12.17} & \textbf{86.75}{\scriptsize$\pm$13.59} & \multicolumn{1}{c|}{\textbf{3.21}{\scriptsize$\pm$10.42}} 
& 91.85{\scriptsize$\pm$11.74} & \textbf{86.90}{\scriptsize$\pm$13.30} & \textbf{3.05}{\scriptsize$\pm$9.76} \\

UMamba-Enc~\cite{maUMambaEnhancingLongrange2024}             
& 91.00{\scriptsize$\pm$13.60} & 85.45{\scriptsize$\pm$16.06} & \multicolumn{1}{c|}{4.11{\scriptsize$\pm$15.09}} 
& \textbf{91.99}{\scriptsize$\pm$11.10} & 86.54{\scriptsize$\pm$13.75} & 3.92{\scriptsize$\pm$15.67} \\

LKM-UNet~\cite{lkmunet}                 
& 90.98{\scriptsize$\pm$13.76} & 85.73{\scriptsize$\pm$16.05} & \multicolumn{1}{c|}{3.79{\scriptsize$\pm$14.15}} 
& 91.34{\scriptsize$\pm$13.03} & 86.14{\scriptsize$\pm$15.29} & 3.55{\scriptsize$\pm$13.47} \\
Diff-UMamba             
& 91.28{\scriptsize$\pm$13.48} & 86.30{\scriptsize$\pm$14.69} & \multicolumn{1}{c|}{3.38{\scriptsize$\pm$10.45}} 
& 91.60{\scriptsize$\pm$11.85} & 86.41{\scriptsize$\pm$14.53} & 3.63{\scriptsize$\pm$14.79} \\ 
\hline
\end{tabular}

\label{tab:brats2}
\end{table}
\noindent In all tasks and datasets, Diff-UMamba shows a consistent increase of 1 to 3\% in IOU compared to UMamba-Bot~\cite{maUMambaEnhancingLongrange2024}. This steady improvement suggests that the Noise Reduction Module (NRM) integrated into Diff-UMamba plays a critical role in enhancing feature representation and model generalization, particularly under the constraints of limited training data. 
\noindent Furthermore, Tables~\ref{tab:brats} and \ref{tab:brats2} show the performance of the sota networks in the BRaTS-21 dataset~\cite{brats1, brats2, brats3} using different proportions of training samples (16\%, 32\%, 48\%, 64\%, 80\%). 
\noindent We compare nnUNetv2~\cite{isenseeNnUNetSelfadaptingFramework2018}, SegResNet~\cite{myronenko3DMRIBrain2018} and mamba architectures~\cite{lkmunet, maUMambaEnhancingLongrange2024} with \mbox{Diff-UMamba}. We observed that Diff-UMamba demonstrated the most substantial improvements when trained on 32\% of the dataset, significantly outperforming other baseline methods in the three metrics (based on a paired t-test with $pvalue < 0.05$).
Due to the presence of multiple input modalities, the difference in dice scores between 16\% and 80\% of the training data is relatively small. Additionally, the benefit of incorporating the NRM is less evident in this setting, as the multimodal input itself helps reduce overfitting.
Thus, the performance improvement was not consistently significant in all smaller dataset sizes (16\% and 48\%).
\begin{figure}[t]
    \centering
    \includegraphics[width=\linewidth]{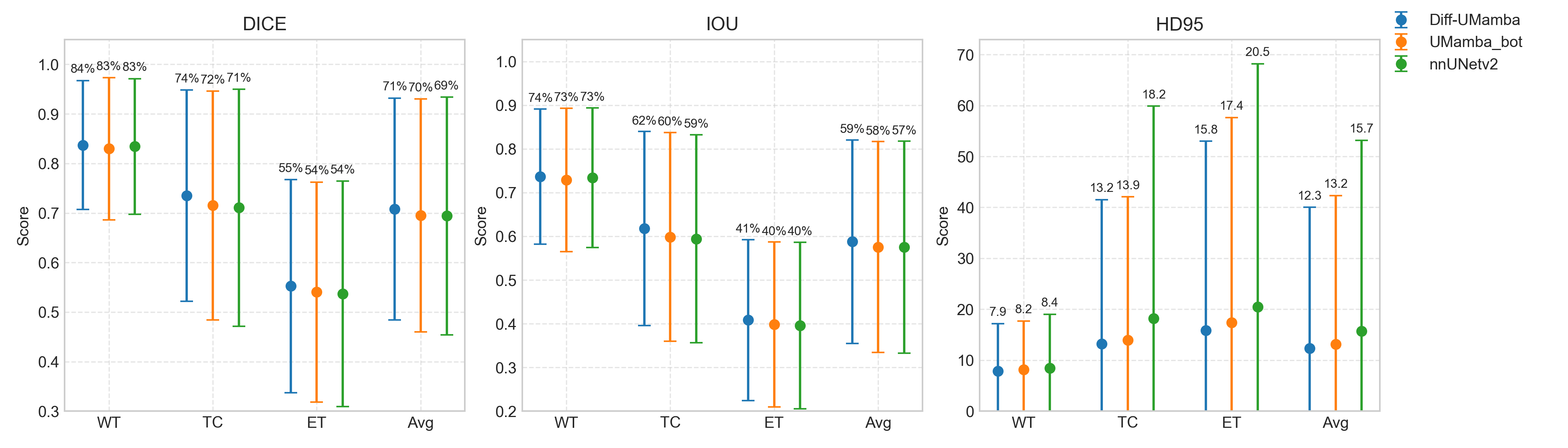}
    \caption{Comparison of DSC, IOU, and HD95 for three models—nnUNetv2~\cite{isenseeNnUNetSelfadaptingFramework2018}, UMamba-Bot~\cite{maUMambaEnhancingLongrange2024}, and Diff-UMamba trained on a single-input T1 weighted MRI. The metrics are reported for whole tumor (WT), tumor core (TC), enhancing tumor (ET), and their average.}
    \label{fig:singlemod}
\end{figure}
Table~\ref{tab:brats2} shows that with larger datasets, UMamba-Bot~\cite{maUMambaEnhancingLongrange2024} and UMamba-Enc~\cite{maUMambaEnhancingLongrange2024} outperform Diff-UMamba and nnUNetv2~\cite{isenseeNnUNetSelfadaptingFramework2018}, reflecting the strength of mamba in modeling long-range dependencies and mitigating the need for NRM.
\noindent This suggests that the benefit of the NRM module is mitigated as the dataset size increases and the base models can generalize more effectively. We repeat the experiment using a single modality, T1-weighted MRI from BraTS-21~\cite{brats1, brats2, brats3}, with 16\% of the training samples, to demonstrate improvements in truly small data settings. Figure~\ref{fig:singlemod} presents a comparison of Diff-UMamba, UMamba-bot~\cite{maUMambaEnhancingLongrange2024}, and nnUNetv2~\cite{isenseeNnUNetSelfadaptingFramework2018} in the three tumor subregions. In particular, Diff-UMamba demonstrates significant improvements in TC, ET and average performance (p<0.05, paired t-test). These results highlight Diff-UMamba’s ability to learn more meaningful representations, particularly in data-constrained environments.

\subsubsection{Ablation Study}

An ablation study was conducted on the internal dataset (using r-CT contours) to evaluate the impact of different configurations of NRM, including the exclusion of $e_1$ to $e_5$, the removal of the mamba block $\mathcal{M}_2$ and variations in the initial value of $\lambda$ ($\lambda_{init}$), as summarized in Table~\ref{tab:ablation}. 
All ablation variants consistently achieved a higher DSC compared to \mbox{UMamba-Bot}.
The optimal configuration (with maximum IOU) was found to include ($e_1$ to $e_5$), retain the mamba block~$\mathcal{M}_2$, and set $\lambda_{init}$ to 0.5. The results indicated that removal of $e_1$ or $e_2$ significantly degraded performance, while excluding deeper layers has a comparatively smaller impact. This supports the hypothesis that the NRM effectively filters noise predominantly captured in the earlier encoder layers. 
\begin{table}[h]
\centering
\setlength{\tabcolsep}{5pt} 
\caption{Ablation study of $m_2$ mamba block, $e_1$ to $e_5$ outputs, 
 and $\lambda_{init}$ value. }
\resizebox{0.6\linewidth}{!}{
\begin{tabular}{cccccccccc}
\hline
\begin{tabular}[c]{@{}c@{}}with $m_2$\\ Mamba\end{tabular} & e1  & e2  & e3  & e4  & e5  & $\lambda_{init}$        & DSC($\uparrow$) & IOU($\uparrow$) & HD95($\downarrow$) \\ \hline

\multicolumn{6}{c}{SegResNet (with NRM)} & - & 73.74 & 63.50 & 3.69 \\

\multicolumn{6}{c}{UNETR (with NRM)} & - & 72.41 & 61.89 & 3.84 \\\hline
\multirow{2}{*}{-}                                                & \multicolumn{5}{c}{\multirow{2}{*}{-}}                          & 0.8                  &      76.67        &  65.47            &   2.99            \\
                                                                  & \multicolumn{5}{c}{}                                                     & 0.1                  &     \textbf{76.97}         &   65.39           &   3.06            \\ \hline
\multirow{5}{*}{-}                                       & $\xmark$  & \cmark & \cmark & \cmark & \cmark &        \multirow{5}{*}{0.5}                       &    74.92          &  64.11            &    3.49           \\
                                                                  & \cmark & $\xmark$  & \cmark & \cmark & \cmark &                               &  73.72            & 62.94             &  3.43             \\
                                                                  & \cmark & \cmark & $\xmark$  & \cmark & \cmark &                               &   76.23           &   65.35           &    3.01           \\
                                                                  & \cmark & \cmark & \cmark & $\xmark$  & \cmark &                               &    75.07          &  63.71            &  3.07             \\
                                                                  & \cmark & \cmark & \cmark & \cmark & $\xmark$  &                              &  76.93             &  65.71            &     \textbf{2.96}          \\\hline
$\xmark$                                                       & \multicolumn{5}{c}{\multirow{1}{*}{-}}                                                     &   \multirow{1}{*}{0.5}                            & 75.28        & 64.12        &  3.11             \\
                                                                  \cmark                                                      & \multicolumn{5}{c}{\multirow{1}{*}{-}}                                   & \multirow{1}{*}{0.5} & 76.91        & \textbf{65.87}        &   3.18            \\
\hline
\end{tabular}}
 \label{tab:ablation}
\end{table}
Furthermore, the results for SegResNet~\cite{myronenko3DMRIBrain2018} and UNETR~\cite{hatamizadehUNETRTransformers3D2021} equipped with the standard NRM module are included in Table~\ref{tab:ablation}, showing modest improvements over their non-NRM counterparts (see Table~\ref{tab:private}), with a dice score gain of 1.01\% for SegResNet and 2.85\% for UNETR. In contrast, Diff-UMamba achieves a significantly higher improvement of 4.44\%, probably because the mamba architecture offers a better inductive bias for integrating the NRM, allowing better modeling of sequential patterns and noise suppression in low-data regimes.

\subsection{Visual Comparison with State-of-the-Art Methods}

Figure~\ref{fig:comp} presents a qualitative comparison between \mbox{Diff-UMamba} and several state-of-the-art segmentation methods in four public datasets: (a) BRaTS-21~\cite{brats1, brats2, brats3}, (b) MSD~\cite{msd} (pancreas), (c) MSD~\cite{msd} (lungs), and (d) AIIB23~\cite{aiib23}. 
\noindent These comparisons highlight the effectiveness of \mbox{Diff-UMamba} in accurately delineating complex anatomical and pathological structures in different modalities and organs.
\noindent In the first example of Figure~\ref{fig:comp}a, Diff-UMamba significantly outperforms other methods in accurately delineating the non-enhancing core (NCR) region. In the second example, while most methods achieve generally similar performance, \mbox{Diff-UMamba} provides more precise delineation in certain subtle regions, as indicated by the red arrows in the figure. In Figure~\ref{fig:comp}b (MSD~\cite{msd}: pancreas), the second example demonstrates a clear advantage of \mbox{Diff-UMamba}: it is the only method that successfully detects the tumor region, whereas most of the other models mistakenly classify it as part of the pancreas. This highlights the sensitivity of the model to small pathological regions and its robustness against class confusion, which is a common issue in pancreatic tumor segmentation due to the subtle appearance and irregular boundaries of the tumor.
\input{visual-comp}
\noindent For the MSD~\cite{msd} (lungs) dataset (Figure~\ref{fig:comp}c), in the first example, most methods exhibit segmentation defects, especially around the boundary region. However, \mbox{Diff-UMamba} shows cleaner and more coherent delineations, with fewer false positives and discontinuities, while LKM-UNet also performs reasonably well compared to other models.
\noindent In the AIIB23~\cite{aiib23} dataset (Figure~\ref{fig:comp}d), which focuses on the segmentation of the airways and surrounding structures, notable artifacts appear in many models in both examples. \mbox{Diff-UMamba}, however, shows greater resilience, yielding more anatomically plausible results with fewer disconnected or fragmented airway segments.
A comparison of the internal dataset is shown in Figure~\ref{fig:internal-comp}, where sub-figures (a) to (c) demonstrate that the incorporation of r-CT contours significantly improves the visibility of tumor boundaries, even in low contrast regions.

\section{Limitations}
\label{sec:limitation}
The noise reduction module (NRM) shows promising results, particularly in small-scale datasets. In this study, NRM was integrated into the UMamba-Bot architecture to leverage the long-range dependency modeling capabilities of mamba on small-scale datasets. 
Alternative integration schemes, such as UMamba-Enc where we can embed NRM into each encoder layer, could enhance denoising performance. However, early encoder layers typically operate on high-resolution feature maps with a large number of tokens, making such integration computationally expensive and nontrivial. Addressing this challenge will be essential in future work to enable a more extensive use of NRM throughout the architecture.
Furthermore, this work is limited to 3D medical image segmentation tasks. The applicability of NRM to 2D data and non-medical domains has not been assessed. 
More broadly, this work opens up a new direction in approaching learning with small datasets. Traditionally, small datasets are paired with lightweight models to avoid overfitting. However, if noise-aware components such as NRM allow larger models to generalize better even in data-scarce settings, this could prompt a rethinking of model design strategies. Instead of simply reducing model size, we might begin exploring larger architectures that are selectively regularized, paving the way for more expressive models even under constrained data regimes. Another scope for future work involves the development of dynamic NRM, which could be activated selectively based on the characteristics of the dataset. For example, NRM could be applied only when a learned noise threshold (e.g., $\lambda_1$ or $\lambda_2$ value) exceeds a certain level, allowing the model to adaptively decide when denoising is necessary. Such a mechanism would enable the dynamic deployment of NRM, making Diff-UMamba more scalable.

\section{Conclusion}
\label{sec:conclusion}
In this work, we introduced Diff-UMamba, a novel model that enhances sequence models such as mamba on smaller datasets by detecting and removing noise from each encoder layer. Our approach is based on \mbox{UMamba-Bot}, integrating a noise reduction module (NRM) to improve performance in limited-data setting.
We demonstrated that it surpasses existing sota models, delivering superior results on both an internal dataset and the public datasets, with limited training samples.
Our findings indicate that Diff-UMamba is suitable for medical imaging applications, where the challenge of obtaining a large number of high-quality annotations remains a significant limitation.

\section*{Acknowledgement}

\small This work was supported by the MINMACS Région Normandie excellence label and ANR LabCom L-Lisa ANR-20-LCV1-0009.
 We thank our colleagues at CRIANN for providing us with the computational resources necessary for this project. We also thank Su Ruan and Tsiry Mayet for proof reading.

 \section*{Disclosure of Interests}{\small The authors have no competing interests to declare that are relevant to the content of this article.}

\bibliographystyle{elsarticle-num}
\bibliography{cas-refs}
\newpage
\appendix
\section{Extra Information of Experiments}
\label{app1}

\begin{figure}[h]
\noindent
\centering

\resizebox{0.70\linewidth}{!}{%
\begin{tikzpicture}[scale = 1]
\input{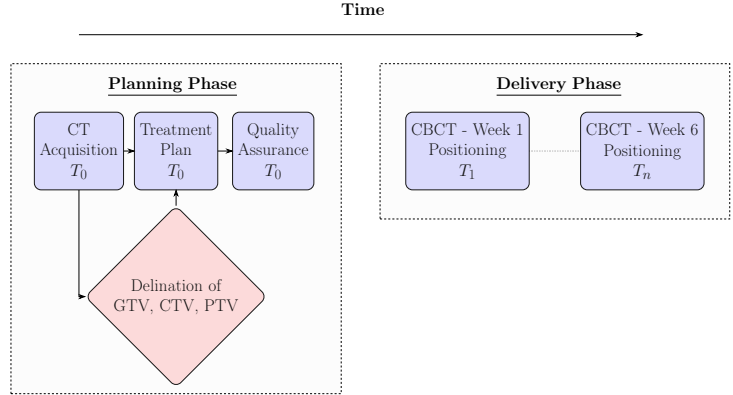}

\end{tikzpicture}
}
\caption{Traditional radiotherapy showing the planning and the delivery phase. It takes around 30-45 days for the treatment during which CBCT images are acquired for positioning.  }

\label{fig:trad-radio}

\end{figure}

\begin{figure}[htbp]
   \centering
 \subfloat[ ]{    \includegraphics[width=0.45\linewidth]{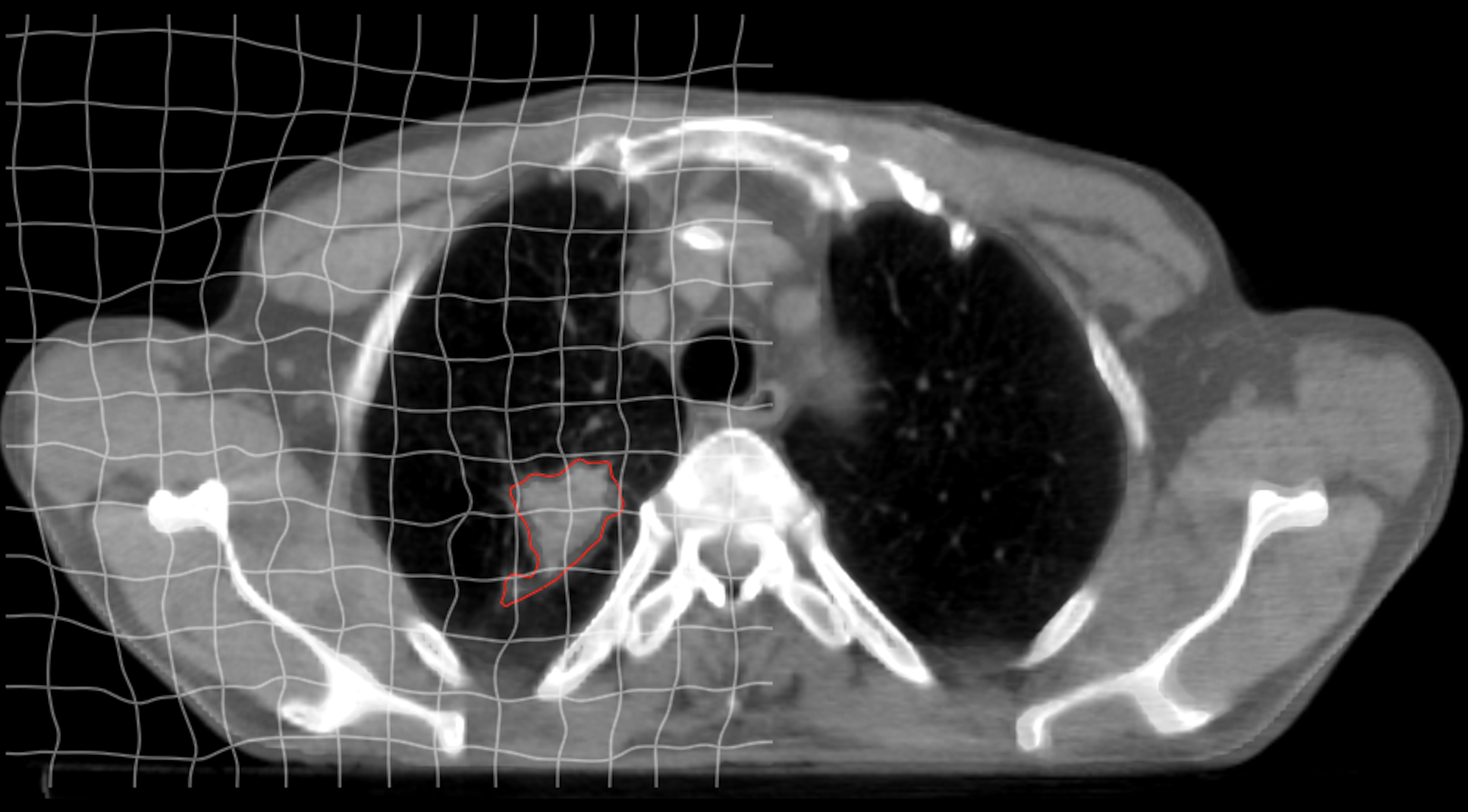}}
 \subfloat[ ]{    \includegraphics[width=0.45\linewidth]{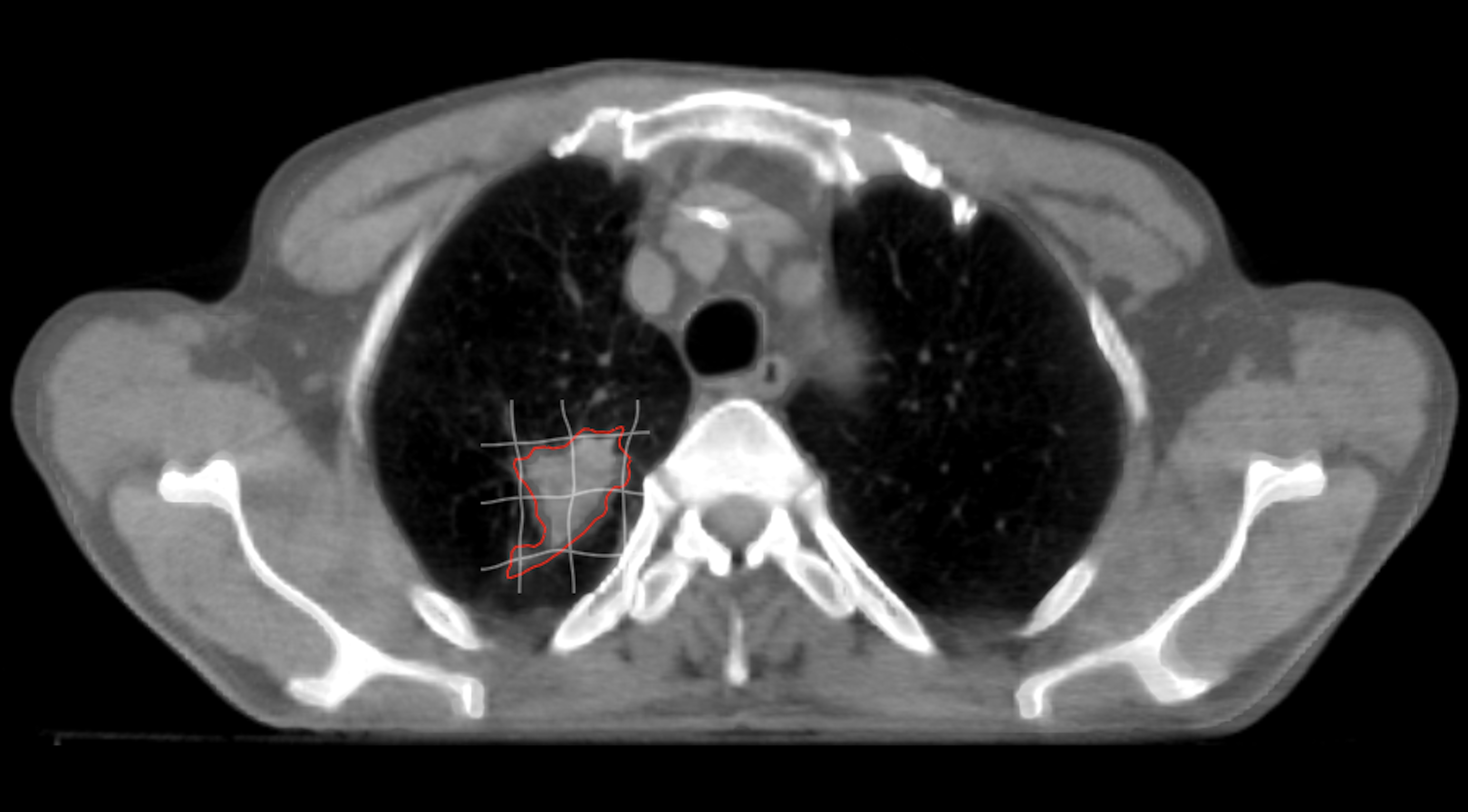}}
 \caption{Deformable vector field representations for two types of deformation, a.) DIR\_L utilizes the lung where the tumor is present, b.) while DIR\_T uses a bounding box to deform only the area around the tumor.}
 \label{fig:dir}
\end{figure}

% \usepackage{multirow}
\begin{table}[H]
\resizebox{\linewidth}{!}{
\begin{tabular}{c|ccc|ccc|ccc|ccc}
\hline
\multirow{2}{*}{Noise Level} & \multicolumn{3}{c|}{Normal}                                                   & \multicolumn{3}{c|}{Speckle}                                               & \multicolumn{3}{c|}{Periodic}                                                  & \multicolumn{3}{c}{Salt and Pepper}                                             \\ \cline{2-13} 
                             & Low/High & Diff-UMamba & UMamba-Bot & Scale & Diff-UMamba & UMamba-Bot & Amplitude & Diff-UMamba & UMamba-Bot & Probability & Diff-UMamba & UMamba-Bot \\ \hline
Level 1                      & 0        & 76.91       & 72.64      & 0     & 76.91       & 72.64      & 0         & 76.91       & 72.64      & 0          & 76.91       & 72.64      \\
Level 2                      & -2/2     & 76.83       & 70.18      & 0.3   & 76.81       & 73.04      & 0.5       & 76.98       & 71.20      & 0.002      & 74.94       & 15.27      \\
Level 3                      & -5/5     & 68.82       & 60.52      & 0.5   & 76.39       & 68.88      & 1         & 76.27       & 66.80      & 0.005      & 68.47       & 0.91       \\
Level 4                      & -8/8     & 64.25       & 29.53      & 0.7   & 74.78       & 59.32      & 2         & 70.77       & 20.18      & 0.008      & 63.47       & 0.69       \\
Level 5                      & -10/10   & 55.69       & 17.45      & 0.9   & 71.05       & 30.86      & 3.5       & 65.55       & 0.073      & 0.01       & 56.79       & 0.56       \\
Level 6                      & -12/12   & 35.25       & 12.36      & 1.1   & 66.03       & 0.07       & 5         & 56.96       & 0.036      & 0.02       & 25.34       & 0.25       \\ \hline
\end{tabular}}
\caption{Noise mapping table corresponding to Figure~\ref{fig:noise}, illustrating the impact of different noise types and intensity to DSC.}
\label{fig:noise-mapping}
\end{table}

\input{demo}

\end{document}